\documentclass[runningheads]{llncs}
\usepackage{graphicx}
\usepackage{graphicx}
\usepackage[font=small,labelfont=bf]{caption}
\usepackage{booktabs}
\usepackage{diagbox}
\usepackage{bm}
\usepackage{algorithm,algpseudocode}
\usepackage[dvipsnames,svgnames,x11names]{xcolor}
\usepackage{amsmath,amssymb} 

\usepackage{mathtools, nccmath}

\usepackage{multirow}
\usepackage[capitalize]{cleveref}
\usepackage{enumitem}
\crefname{section}{Sec.}{Secs.}
\Crefname{section}{Section}{Sections}
\Crefname{table}{Table}{Tables}
\crefname{table}{Tab.}{Tabs.}

\RequirePackage{cite}     

\usepackage{float}
\floatstyle{plaintop}
\restylefloat{table}

\usepackage{subcaption}
\usepackage{tikz}
\usepackage{comment}
\usepackage{color}

\usepackage[accsupp]{accessibility}  

\makeatletter
\DeclareRobustCommand\onedot{\futurelet\@let@token\@onedot}
\def\@onedot{\ifx\@let@token.\else.\null\fi\xspace}

\makeatother

\newcommand{\argmin}{\operatornamewithlimits{argmin}}

\RequirePackage{silence}  

\definecolor{darkgreen}{RGB}{30,150,30}
\definecolor{darkblue}{RGB}{0,0,127}
\definecolor{darkyellow}{RGB}{171,133,0}
\definecolor{darkred}{RGB}{180,20,20}
\definecolor{darkmagenta}{RGB}{127,0,127}
\definecolor{darkcyan}{RGB}{0,127,127}
\definecolor{orange}{rgb}{0.8, 0.4, 0.0}

\newif\ifdrafting 
\draftingtrue 

\ifdrafting
  \newcommand{\VZ} [1] {\textcolor{darkgreen}{[VZ: #1]}}
  \newcommand{\AK} [1] {\textcolor{darkred}{[AK: #1]}}
  \newcommand{\RD} [1] {\textcolor{BurntOrange}{[RD: #1]}}
  
  \newcommand{\EL} [1] {\textcolor{darkmagenta}{[EL: #1]}}
  \newcommand{\CG} [1] {\textcolor{darkmagenta}{[CG: #1]}}
  \newcommand{\TODO} [1] {{\color{darkcyan}{\bf [TODO: #1]}}}
  
\else
  \newcommand{\VZ} [1] {}
  \newcommand{\AK} [1] {}
  \newcommand{\RD} [1] {}
  \newcommand{\EL} [1] {}
  \newcommand{\CG} [1] {}
  \newcommand{\TODO} [1] {}
  
\fi

\begin{document}

\pagestyle{headings}
\mainmatter
\def\ECCVSubNumber{5257}  

\title{Learned Monocular Depth Priors in Visual-Inertial Initialization} 

\titlerunning{Learned Monocular Depth Priors in Visual-Inertial Initialization}
%
\author{Yunwen Zhou \and Abhishek Kar \and Eric Turner \and Adarsh Kowdle \and Chao X. Guo \and Ryan C. DuToit \and Konstantine Tsotsos}
\authorrunning{Y. Zhou et al.}
%
\institute{Google AR
\email{\{verse,abhiskar,elturner,adarshkowdle,chaoguo,rdutoit,ktsotsos\}@google.com}}
\maketitle
\begin{sloppypar}

\begin{abstract}
Visual-inertial odometry (VIO) is the pose estimation backbone for most AR/VR and autonomous robotic systems today, in both academia and industry. However, these systems are highly sensitive to the initialization of key parameters such as sensor biases, gravity direction, and metric scale. In practical scenarios where high-parallax or variable acceleration assumptions are rarely met (e.g. hovering aerial robot, smartphone AR user not gesticulating with phone), classical visual-inertial initialization formulations often become ill-conditioned and/or fail to meaningfully converge. In this paper we target visual-inertial initialization specifically for these low-excitation scenarios critical to in-the-wild usage. We propose to circumvent the limitations of classical visual-inertial structure-from-motion (SfM) initialization by incorporating a new learning-based measurement as a higher-level input. We leverage learned monocular depth images (mono-depth) to constrain the relative depth of features, and upgrade the mono-depths to metric scale by jointly optimizing for their scales and shifts. Our experiments show a significant improvement in problem conditioning compared to a classical formulation for visual-inertial initialization, and demonstrate significant accuracy and robustness improvements relative to the state-of-the-art on public benchmarks, particularly under low-excitation scenarios. We further extend this improvement to implementation within an existing odometry system to illustrate the impact of our improved initialization method on resulting tracking trajectories.
\end{abstract}

\section{Introduction}
\label{sec:intro}
\begin{figure*}[!ht]
    \centering
    \begin{subfigure}{1.\linewidth}
    \centering
    \includegraphics[width=0.8\linewidth]{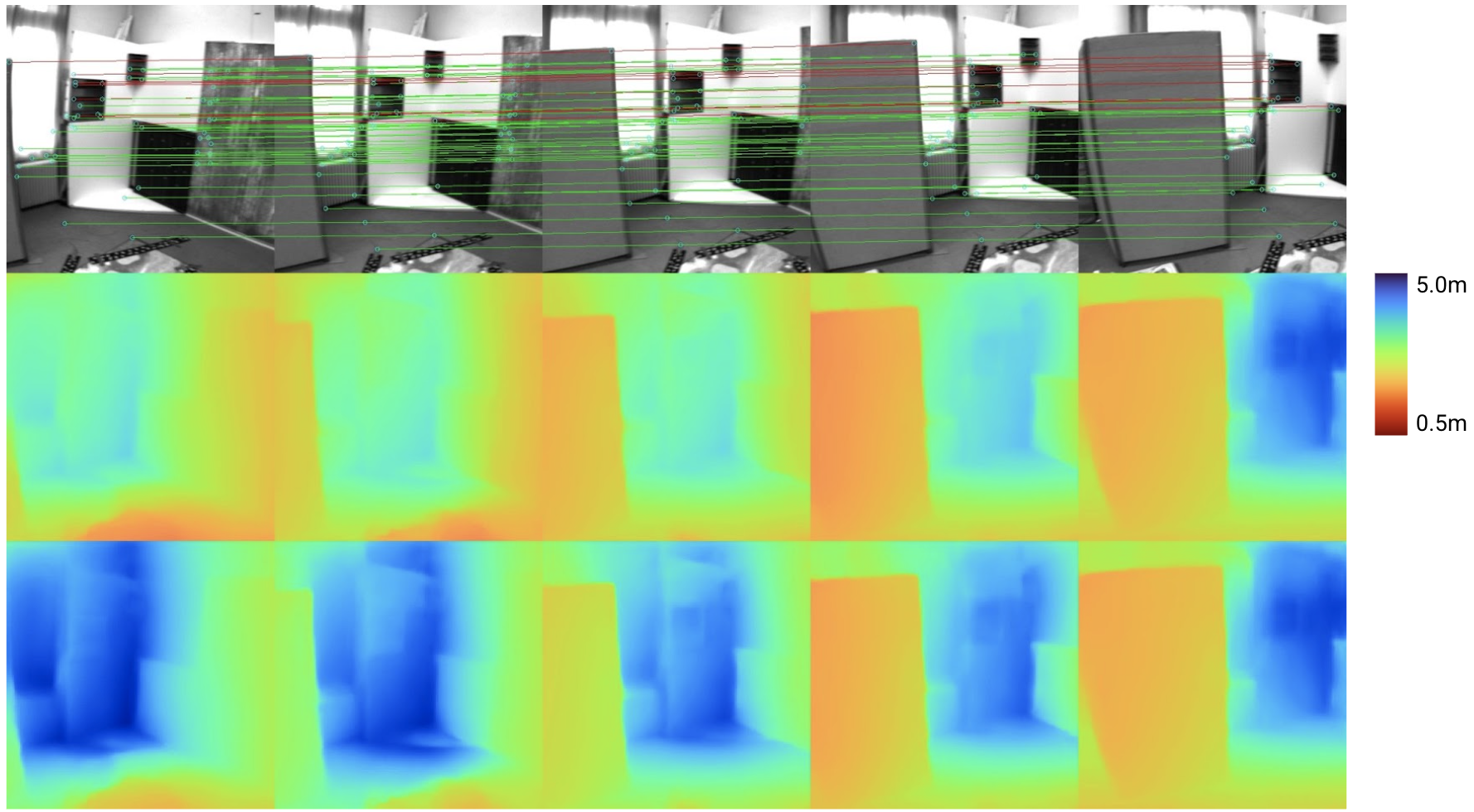}
    \caption{\textbf{First Row:} Intensity image inputs. \textbf{Second Row:} Mono-depth images. \textbf{Third Row:} Metric-depth images, recovered after joint motion, scale, and shift optimization. Stable metric-depth is recovered after the optimization from initial inconsistent and inaccurate mono-depth. \textbf{Green Tracks on First Row:} Inlier feature-tracks for mono depth constraints. \textbf{Red Tracks on First Row:} Outlier feature-tracks due to temporally inconsistent associated mono-depth values (see \cref{depth constraint})}
    \label{depth constraint vis}
    \end{subfigure}
    \vfill
    \begin{subfigure}{\textwidth}
    \centering
    \includegraphics[width=0.7\linewidth]{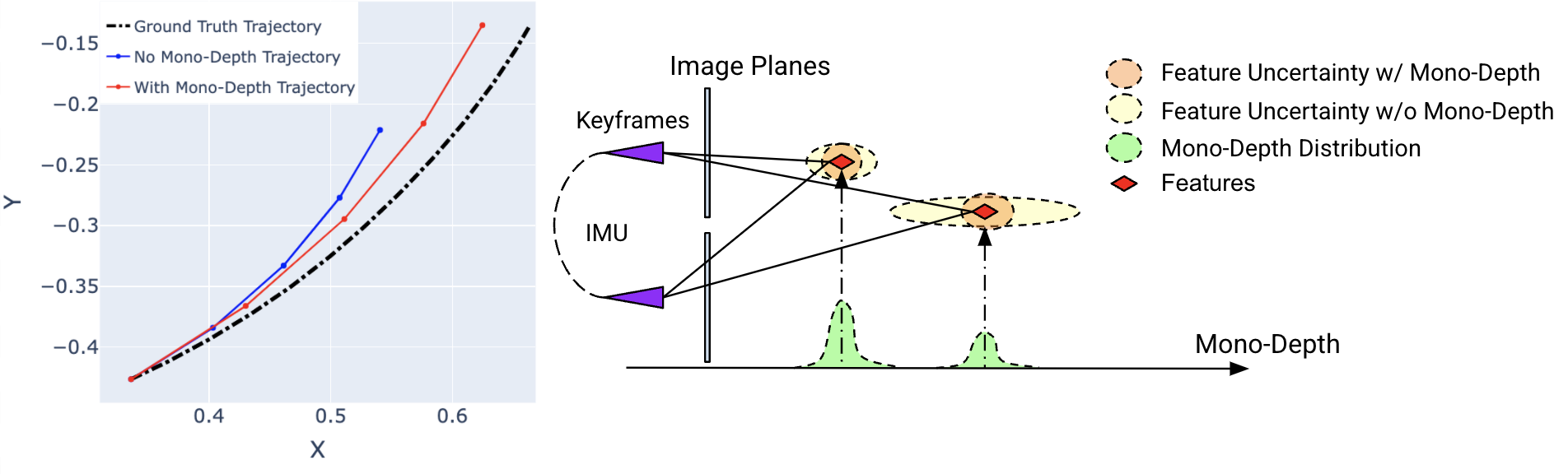}
    \caption{\textbf{Left}: Initialization trajectory under a limited motion scenario in meters. Trajectory recovery is improved with tight coupling between VI-SFM and mono-depth (note incorrect scale in blue trajectory). \textbf{Right}: Mono-depth coupling improves problem conditioning, potentially reducing uncertainty of estimates and increasing accuracy.}
    \label{uncertainty vis}
    \end{subfigure}
    \caption{At top, demonstration of depth constraints over an initialization window. At bottom, demonstration of trajectories estimated with and without mono depth on the sequence shown at \textbf{top}, illustration of feature position uncertainty.}
\end{figure*}

Monocular visual-inertial odometry (VIO) enables accurate tracking of metric 3D position and orientation (pose) using just a monocular camera and inertial measurement unit (IMU) providing linear acceleration and rotational velocity. These techniques have unlocked an economical and near-ubiquitous solution for powering complex scene understanding in augmented or virtual reality (AR/VR) experiences (e.g. \cite{depthlab}) on commodity platforms (e.g, Google's ARCore and Apple's ARKit), alongside other robotic applications such as aerial delivery drones. A precondition of successful operation in these scenarios is successful (and accurate) initialization of key system parameters such as scale, initial velocity, accelerometer and gyro biases, and initial gravity direction. Poor initialization typically leads to tracking divergence, unacceptable transients, low-accuracy operation, or outright failures, especially of downstream modules (e.g. drone navigation software). Unfortunately, visual-inertial initialization routines have a very common failure mode in these realistic scenarios: insufficient motion for the system's motion and calibration states to be unambiguously resolvable ~\cite{jones2007inertial, vins_on_wheels, josh_observability, martinelli2014closed, kelly2011visual}. This occurs, for example, if the user of a phone-based AR game moves with very little parallax relative to the visible scene or when a drone must initialize while hovering. These are extremely common in practice. To improve VIO initialization in these scenarios on commodity hardware we must optimize for the total (user-visible) latency to initialization and accuracy of the resulting trajectories, while not violating real-time operation. For example, a phone-based AR user may expect a responsive ($<500ms$) startup of their game, regardless of how they moved their phone, and without taking noticeable compute resources from the primary AR application. 

Due to its impact, many recent works have focused on formulating fast and accurate initialization algorithms for robust monocular VIO~\cite{Inertial-only, analytical-solution, vins-mono-initialization, convex-initialization, martinelli2014closed, gyro-bias-calibration}. These works rely on sparse visual feature tracks to constrain relative pose (up to scale) in the visual-inertial structure-from-motion (VI-SFM) problem. Under low parallax initialization scenarios, any classical depth estimation approach for these features in the VI-SFM problem will be susceptible to large uncertainty, such as in the sequence in \cref{depth constraint vis}. This uncertainty (illustrated in \cref{uncertainty vis}) makes the overall system ill-conditioned, often resulting in poor or failed initializations. This ambiguity is exacerbated if the inertial measurements lack enough variation to reliably recover metric scale \cite{martinelli2014closed}.


Inspired by the robustness achievements of depth-enabled visual SLAM systems~\cite{orb-slam2, rgbd-slam, RGBDTAM, elasticfusion} and recent advances in generalized learning-based monocular depth (mono-depth) \cite{midas, intel-dpt}, we propose a novel formulation of monocular VIO initialization. We incorporate depth measurements from a mono-depth model directly into a classical VI-SFM framework as measurements. Our proposed method operates in real-time on a mobile phone and is able to accurately initialize in traditionally challenging low parallax or limited acceleration scenarios, \textit{without} requiring an additional dedicated sensor for estimating depth (e.g. LiDAR, Time-of-Flight). Our primary contributions are:
\begin{itemize}
    \setlength\itemsep{0.1em}
    \item We apply learned monocular depth priors for VIO initialization. To the best of our knowledge, we are the first to leverage the power of learned depth for this problem through coupling with classical methods.
    \item We propose a novel residual function which tightly couples scale and shift invariant monocular depth measurements within a traditional VI-SFM formulation.
    \item We propose a gradient-based residual weighting function and an outlier rejection module to effectively deal with noisy depth predictions.
    \item We demonstrate robust and accurate initialization relative to the state-of-the-art on public benchmarks when embedded within an existing tracking system, particularly under low-excitation scenarios (i.e. when accelerometer readings or velocity do not significantly change across the initialization window). We achieve all of the above while maintaining real-time performance on 10Hz image streams on resource constrained devices.
\end{itemize}

\section{Related Work}
\label{sec:related_work}
Visual-inertial odometry~\cite{vio-tutorial, vins-review} is a well-studied problem in both the computer vision and robotics communities and many works~\cite{msckf, vins_mono, tsotsos2015robust, orb-slam3, svo, dso-imu, fei2019xivo, rovio, okvis} have focused specifically on accurate initial estimation of states required by the inertial sensor. These works can be roughly classified into two categories - 1) jointly solving a visual-inertial SFM problem directly in closed form or as a bundle adjustment problem~\cite{martinelli2014closed, MK+BA-solution, convex-initialization} and 2) cascaded approaches which solve a pure visual SFM for up to scale pose followed by metric scale recovery using inertial observations~\cite{analytical-solution, Inertial-only, vins-mono-initialization, vertical-edge-vio-initialization}. Both approaches typically use a visual-inertial bundle adjustment (VI-BA) step to further refine their solution.

Feature-based visual odometry (VO) plays a key role in VIO initialization but often exhibits large uncertainty in low parallax and motion scenarios. Additionally, the VO prior requires enough non-zero inertial measurements for observing metric scale~\cite{martinelli2014closed} to initialize VIO. A recent state-of-the-art method~\cite{Inertial-only} (used as the initialization routine for the popular ORBSLAM3 system~\cite{orb-slam3})  still requires around $2$ seconds (at 10Hz) to initialize and only succeeds with reasonable motion excitation. Our proposed method aims to initialize with lower (user-visible) latency (i.e. less data collection time) even in challenging low-motion scenarios. Some prior works have explored using higher order visual information such as lines~\cite{vertical-edge-vio-initialization} for increased system observability in monocular VIO. Additionally, RGB-D SLAM systems~\cite{orb-slam2, rgbd-slam, RGBDTAM} have been tremendously successful in a number of domains (AR/VR, self driving cars, etc.) and can inherently initialize faster given direct metric depth observations. For example,~\cite{RGBD-extrinsics} demonstrated that the inclusion of a depth sensor significantly reduces the required number of feature observations. However, in spite of their advantages, depth sensors can significantly increase the cost and/or complexity of a device. Our work is focused on improving VIO initialization for commodity devices equipped with only an IMU and single camera.

With the advent of deep learning, there has been significant interest in end-to-end learning for VIO~\cite{deepvio, deepvo, vinet, selfvio, selective-VIO, deepvo+ekf}. However, the proposed methods often lack the explainability and modular nature of traditional VIO systems, have alternative end-goals (e.g. self supervised depth/optical flow/camera pose estimation), or are too expensive to operate on commodity hardware without custom accelerators. Moreover, end-to-end methods don't explicitly consider in-motion initialization and often benchmark on datasets with the trajectory starting at stationary point ~\cite{kitti, euroc}. Prior works have also explored learning methods in purely inertial~\cite{liu2020tlio,ronin-io,chen2018ionet} or visual systems~\cite{bloesch2018codeslam,tang2018ba,kopf2021rcvd}. CodeVIO~\cite{codevio} demonstrated that incorporating a differentiable depth decoder into an existing VIO system (OpenVINS) ~\cite{openvins} can improve tracking odometry accuracy. Note that CodeVIO does not tackle the VIO initialization problem and relies on tracking landmarks from already-initialized VIO. It uses the OpenVINS initialization solution which only initializes after observing enough IMU excitation following a static period. However, CodeVIO does demonstrate an effective and modular integration of learned priors within VIO and inspires us to deliver similar improvements to VIO initialization, while operating under realtime performance constraints.

\section{Methodology}
\begin{figure}[t]
    \centering
    \includegraphics[width=\linewidth]{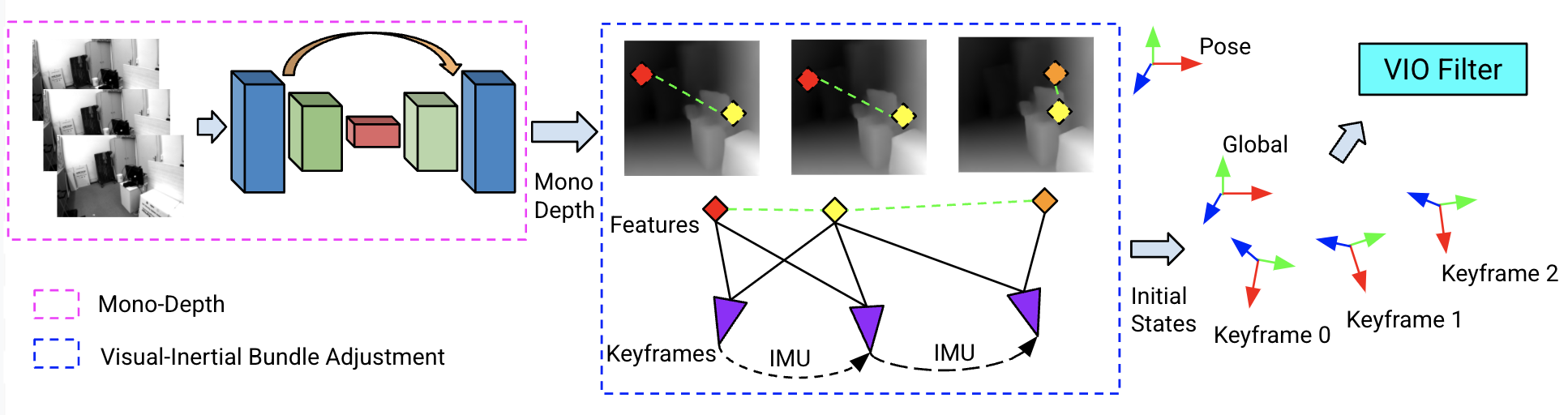}
    
    \caption{Overall initialization diagram composed of monocular depth inference module running on each keyframe, and the visual-inertial bundle adjustment module. Initialized states are then fed into our VIO for tracking.}
    \label{fig:system diagram}
\end{figure}

Our proposed system is composed of two modules as shown in \cref{fig:system diagram}: 1) monocular depth inference which infers (relative) depth from each RGB keyframe, and 2) a VIO initialization module which forms a visual-inertial structure-from-motion (VI-SFM) problem, with the relative depth constraints from the inferred monocular depth. This VI-SFM problem aims to estimate keyframe poses, velocity, and calibration states, which are then used as the initial condition for a full VIO system.

Like most VIO initialization algorithms \cite{MK+BA-solution, analytical-solution, Inertial-only}, our VIO initialization consists of a closed-form solver, whose solution is then refined with visual-inertial bundle adjustment (VI-BA). In this section, we first briefly describe our mono-depth model. Then, we detail our contribution on employing mono-depth constraints in VI-BA refinement.

\subsection{Light-weight Monocular Depth Model}

\label{mono-depth model}
Our key contribution in this work is to incorporate prior-driven monocular depth constraints within a classical VIO initialization framework for better tracking initialization. For the final system to be practical, we require the mono-depth model to generalize to a wide variety of scenes and operate under a small compute budget. We follow recent state-of-the-art monocular depth estimation models~\cite{midas} and train a lightweight mono-depth network. Specifically, we use the robust scale-shift invariant loss~\cite{midas} alongside various edge-sensitive depth losses~\cite{midas,  mannequin_challenge} and train a small UNet model on a variety of datasets including ScanNet~\cite{dai2017scannet}, MannequinChallenge~\cite{mannequin_challenge} as well as pseudo-ground truth disparity maps generated on the OpenImages~\cite{OpenImages2} dataset using large pretrained publicly available models~\cite{midas}. For datasets with metric depth ground truth (e.g. ScanNet), we also add a loose metric depth loss term (Charbonnier loss~\cite{barron2019general} between prediction and inverse metric depth) to inform the scale and shift priors in \cref{scale-shift-prior}. We trained our model on gravity-aligned (or ``upright'') images to avoid having it learn depth maps for ``sideways'' images and better use its limited model capacity. Our final model is fast (\cref{computation-cost}), light-weight ($\sim600K$ parameters) and predicts relative (inverse) depth maps as shown in \cref{depth constraint vis}.

Given the scale-shift invariant nature of our training losses, the metric inverse depth, $z$, can be expressed as a scaled and shifted version of the model prediction, $d$, as $z = ad + b$, where $a$ and $b$ are the scale and shift parameters respectively. Moreover, as our model is trained on gravity aligned (``upright") images, we rotate the input image in 90-degree increments before inferring depth. Since only 45-degree accuracy is required to get the best rotation, for simplicity we use accelerometer measurements rotated through pre-calibrated IMU-camera extrinsics as an estimate of gravity in the camera frame.

\subsection{VI-BA with Monocular Depth Constraints}
\label{BA in methodology}
We aim to solve for the following state parameters, $\bm{\mathcal{X}}$, in our VI-BA problem
\begin{equation}
\label{ba X}
    \bm{\mathcal{X}} = [\bm{X_0}; \dots; \bm{X_{N-1}}; \bm{{}^{C_j}f_{0}}; \dots; \bm{{}^{C_j}f_{M-1}}; \bm{S_0}; \dots; \bm{S_{N-1}}]
\end{equation}
\label{estimate value explanation}
where 
\begin{itemize}
    \setlength\itemsep{0.1em}
    \item $\bm{X_k}$ represents the $k^{th}$ IMU keyframe state among $N$ keyframes in total, which is $[\bm{q_{k}}; \bm{p_{k}}; \bm{v_{k}}; \bm{b^a_{k}}; \bm{b^\omega_{k}}]$. $\bm{q_{k}}$ and $\bm{p_{k}}$ are the $k^{th}$ IMU keyframe pose parameterized as quarternion and translation w.r.t the global frame $\{G\}$ in which we assume the direction of gravity is known. $\bm{v_{k}}$ is the velocity in $\{G\}$ and $\bm{b^a_{k}}$, $\bm{b^\omega_{k}}$ are the accelerometer and gyro biases at the $k^{th}$ keyframes. 
    \item $\bm{{}^{C_j}f_{i}}$ represents the $i^{th}$ feature point parameterized in local inverse depth $[u_{ij}, v_{ij}, w_{ij}]^T$ with respect to the $j^{th}$ keyframe's camera coordinates. $u_{ij}$ and $v_{ij}$ lie on normalized image $XY$ plane and $w_{ij}$ is the inverse depth \cite{inversedepth}.
    \item $\bm{S_k}  = [a_k; b_k] $ following \cref{mono-depth model}, which are scale and shift for recovering metric depth from the raw mono-depth at the $k^{th}$ keyframe. 
    \item The IMU-camera extrinsics ($\bm{q_C}$, $\bm{p_C}$) and 3D-2D projection parameters $Proj(\cdot)$ are not estimated due to lack of information in such a small initialization window. We adopt pre-calibrated values as is customary.
\end{itemize}

We initialize the state $\bm{\mathcal{X}}$ using a standard closed-form solver~\cite{convex-initialization} for a VI-SFM problem formulated with reprojection error. Its formulation and derivation are presented in the supplemental material. Given keyframes $\mathcal{K}$, with up to scale and shift mono inverse depth, feature points $\mathcal{F}$, and $\mathcal{L} (\subset\mathcal{F})$ feature points with mono inverse depth measurements, the VI-BA minimizes the following objective function:

\begin{equation}
\label{ba_objective_function}
\begin{split}
\bm{\hat{\mathcal{X}}} = &\argmin_{\bm{\mathcal{X}}} \underbrace{\sum_{(i, j)\in \mathcal{K}} \|\bm{r_{\mathcal{I}_{ij}}}\|^2_{\Sigma_{ij}}}_{\text{Inertial\ Constraints}} + \underbrace{\sum_{i\in\mathcal{F}}\sum_{k\in\mathcal{K}}\rho(\|\bm{r_{\mathcal{F}_{ik}}}\|^2_{\Sigma_{\mathcal{F}}})}_{\text{Visual\ Constraints}} \\
&+ \underbrace{\sum_{i\in\mathcal{L}}\sum_{k\in\mathcal{K}}\lambda_{ik} \rho(\|r_{\mathcal{L}_{ik}}\|^2)}_{\text{\textbf{Mono-Depth\ Constraints}}} + \underbrace{\|\bm{r_0}\|^2_{\Sigma_0} + \sum_{i \in \mathcal{K}}\|\bm{r_{\mathcal{S}_i}}\|^2_{\Sigma_\mathcal{S}}}_{\text{Prior\ Constraints}}
\end{split}
\end{equation}
where $\bm{r_{\mathcal{I}_{ij}}}$ is the IMU preintegration residual error~\cite{preintegration} corresponding to IMU measurements between two consecutive keyframes, $\bm{r_{\mathcal{F}_{ik}}}$ is the standard visual reprojection residual resulting from subtracting a feature-point's pixel measurement from the projection of $f_i$ into the $k^{th}$ keyframe~\cite{multiview_geometry}, $\bm{r}_{\mathcal{L}_{ik}}$ is an inverse depth temporal consistency residual for incorporating mono-depth, and $\bm{r_{\mathcal{S}_i}}$ is a residual relative to a prior for scale and shift (\cref{depth constraint}). $\bm{r}_0$ is a prior for the bias estimates of the $0th$ keyframe and $\Sigma_0$, $\Sigma_{ij}$, $\Sigma_{\mathcal{F}}$, $\Sigma_{\mathcal{S}}$ are the corresponding measurement covariance matrices. $\lambda_{ik}$ is a scalar weight for each depth residual and $\rho(.)$ refers the huber-loss function~\cite{huber-loss}.

\begin{figure}[t]
    \centering
    \includegraphics[width=0.65\linewidth]{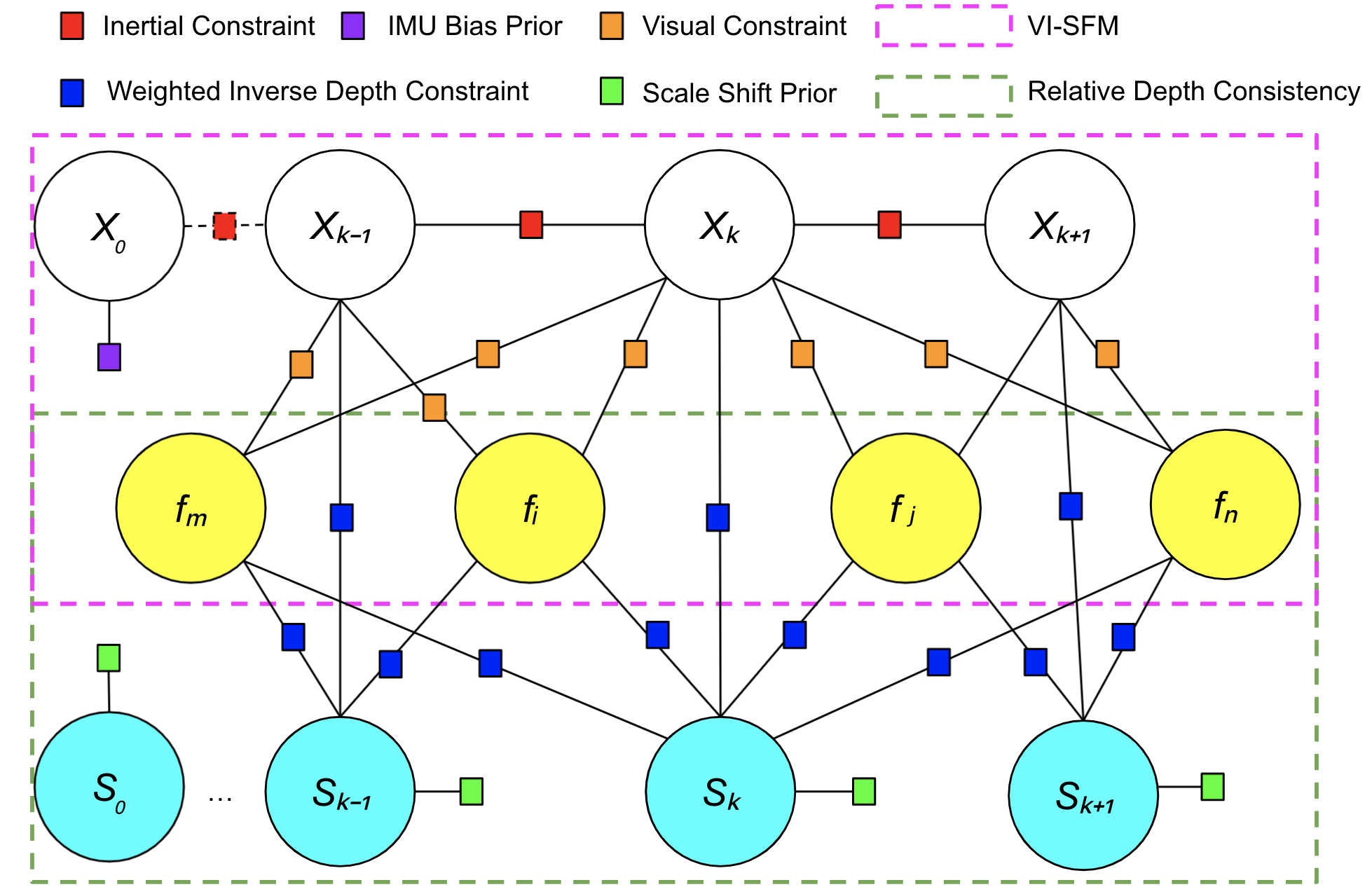}
    \caption{A factor graph illustration of the VI-SFM depth refinement problem \cref{ba_objective_function}. Circled nodes represent $\bm{\mathcal{X}}$ in \cref{ba X} to be estimated. They are connected by constraints illustrated in the graph. The \textbf{pink dashed box} is the traditional VI-SFM problem. The \textbf{green dashed box} represents the new proposed constraints to maintain relative feature depth consistency across keyframes. Feature points and poses are constrained through the scale-shift parameters $\bm{S}$.}
    \label{fig:Bundle Adjustment Factor Graph}
\end{figure}

The factor graph resulting from \eqref{ba_objective_function} is illustrated in \cref{fig:Bundle Adjustment Factor Graph}. ($\bm{r_{\mathcal{I}_{ij}}}$, $\bm{r_{\mathcal{F}_{ik}}}, \bm{r}_0$) forms the traditional VI-SFM problem as highlighted in the pink dashed box. The following sections detail the proposed depth constraints ($r_{\mathcal{L}_{ik}}, \bm{r_{\mathcal{S}_i}}$) which are grouped by green dashed box. 

\subsection{Weighted Mono-Depth Constraints}
\label{depth constraint}

As illustrated in \cref{fig:Bundle Adjustment Factor Graph}, depth constraints relate observed feature-point depth with that keyframe's scale-shift parameters, $\bm{S_k}$. Hence only $2$ additional parameters are needed to model the hundreds of mono-depth residual equations for each keyframe-landmark pair. As demonstrated in \cref{sec:experiments}, this improves the system conditioning under motion restricted scenarios.

The depth constraints comprise three major components - the \textbf{residual function}, the \textbf{weight} for each residual and the \textbf{outlier rejection} module to reject inconsistent mono-depth measurements across keyframes.

\textbf{Inverse Depth Residual Function.} Inspired by the loss functions employed in monocular deep depth estimation \cite{mono_ml_depth}, our proposed depth residual for keyframe $k$ and feature point $i$ takes the form of the $log$ of the ratio between the measured depth scaled/shifted by $\bm{S_k}$ and the feature point's estimated depth:

\begin{equation}
\label{inverse depth residual}
 r_{\mathcal{L}_{ik}} = \log \left((a_k d_{ik} + b_k) \cdot\Omega(\bm{{}^{C_j}\text{\hspace{-1mm}}f_i}, \bm{q_{j}}, \bm{p_{j}}, \bm{q_{k}}, \bm{p_{k}}\right))
\end{equation}
Where $\Omega(\cdot)$ is the depth of the feature point $i$ (which is parameterized with respect to keyframe $j$) in keyframe $k$. If $k=j$ then $\Omega(\cdot)$ can be simplified to $w^{-1}_{ij}$. This is how we tie mono-depth parameters to multiple features and poses to better constrain the problem. The derivation details for $\Omega(\cdot)$ are presented in supplemental material.

It is well known that this  residual can lead to a degenerate solution of scale going to zero or a negative value \cite{scale-degeneration-ideas}. To avoid this, we adopt the common technique of defining the scale parameter $a_k$ as
\begin{equation}
    a_k = \varepsilon + log(e^{s_k} + 1)
\end{equation}
where $\varepsilon = 10^{-5}$, which prevents $a_k$ from being either negative or zero, allowing us to optimize  $s_k$ freely.

\textbf{Scale-shift Prior.} Reiterating \cref{mono-depth model}, the ML model is trained on certain metric depth datasets with a loss where the scale is supposed to be $1$ and shift is $0$. We define prior residuals for scale and shift at the $i^{th}$ frame as
\begin{equation}
\label{scale-shift-prior}
    \bm{r}_{\mathcal{S}_i} = \begin{bmatrix} 1 - a_i && -b_i \end{bmatrix}^T
\end{equation}
Since metric depth is not observable from the ML model, in practice we assign a \emph{very} large covariance $\Sigma_\mathcal{S}$ to these scale-shift priors terms ($0.3$ for scale, $0.2$ for shift), which keeps parameters bounded to the regime in which model training occurred, and in degenerate situations such as zero-acceleration, allows us to converge to a sensible scale. 

\cref{depth constraint vis} shows the effectiveness of the depth constraints and scale-shift priors. With them, we are able to upgrade the learned depth to metric level. The better-conditioned problem then yields a more accurate trajectory, illustrated in \cref{uncertainty vis}.

\textbf{Edge Awareness Weight.} The ML model doesn't explicitly yield prediction uncertainty, however, we empirically observe the uncertainty is larger near depth edges and propose a loss weight, $\lambda_{ik}$, which modulates the residual with gradients of image $I_k$ and depth $D_k$ as follows

\begin{equation}
\label{weight function}
\begin{split}
&\lambda_{ik} = e^{-(\alpha |\nabla^2 \Phi(I_k(u_{ik}, v_{ik}))| + |\nabla^2 \Phi(D_k(u_{ik}, v_{ik}))|)}
\end{split}
\end{equation}
where $\nabla^2$ is the laplacian operator, $\Phi(\cdot)$ is a bilateral filter for sharpening image and depth edges, $\alpha$ is a hyperparameter for relative weighting of image/depth gradients and $(u_{ik},v_{ik})$ is the pixel location of the feature point in keyframe $k$. This weight diminishes the effect of depth constraints on feature points near image/depth edges and favors non-edge regions where the depth and image gradients are in agreement.

\textbf{Outlier Rejection for Depth Measurements.}
The weighting function \cref{weight function} helps mitigate effects of erroneous mono-depth measurements at a given keyframe, but cannot reconcile inconsistency in depth measurements across keyframes. For a short initialization window ($<2s$), keyframe images tend not to vary drastically. Given this, we expect the mono-depth output to not vary significantly as well (even though they are up to an unknown scale and shift). For example, if the mono-depth model predicts a feature point to have small depth w.r.t the rest of the scene in one keyframe but large depth in another, the mono-depth residuals for this given feature are likely to be unreliable and should not be included in the final optimization.

\begin{center}{
\begin{minipage}{0.7\linewidth}
\begin{algorithm}[H]
\caption{Outlier Depth Measurements Rejection}
\label{outlier rejection}
\begin{algorithmic}[1]
\small
\Require Mono-depth residuals $r_{\mathcal{L}ik}, i \in \mathcal{L}, k \in \mathcal{K}$; thresholds $\sigma_\text{min}, \sigma_\text{max}$
\Ensure Set of inlier mono-depth residuals
\State $\sigma_{\mathcal{L}} \leftarrow \{\}$
\For{$i \in \mathcal{L}$}
    \State Append $\sigma_i = \sqrt{\frac{\sum_k\left(r_{ik} - \hat{r}_i\right)}{N-1}}$ to $\sigma_{\mathcal{L}}$
\EndFor

\If{$\text{percentile}(\sigma_{\mathcal{L}}, 25) > \sigma_{\text{max}}$}

\Return $\{\}$

\ElsIf{$\text{percentile}(\sigma_{\mathcal{L}}, 85) < \sigma_{\text{min}}$}

    \Return $\{r_{\mathcal{L}ik}, \forall i \in \mathcal{L}, \forall k \in \mathcal{K}\}$
\Else

    \Return $
    \{r_{\mathcal{L}ik} | \sigma_i < \text{percentile}(\sigma_{\mathcal{L}}, 85)\}$
\EndIf
\end{algorithmic}
\end{algorithm}
\end{minipage}
\par
}\end{center}

Thus, we devise an outlier-rejection scheme detailed in \cref{outlier rejection}. 
This algorithm first evaluates the standard deviations of residuals involving a given feature point, $\sigma_\mathcal{L} = \left\{\sigma_i, \forall i \in \mathcal{L}\right\}$. Then depending on the distribution of $\sigma_\mathcal{L}$ we choose the inlier set. (i) If the $25^{th}$ percentile of $\sigma_\mathcal{L}$ is larger than a maximum threshold, we reject all mono-depth constraints. This scenario occurs when the ML inference is highly unstable and typically does not yeild useful constraints.  (ii) When mono-depth constraints are generally self-consistent (the $85^{th}$ percentile of $\sigma_\mathcal{L}$ is smaller than a minimum threshold) we accept all mono-depth constraints. (iii) In all other cases, we reject residuals corresponding to $\sigma_i$ in upper $15^{th}$ percentile of $\sigma_\mathcal{L}$, removing the least self-consistent constraints. Such a scenario is depicted in \cref{depth constraint vis}, where the mono-depth residuals involving red feature tracks are rejected.

In practice, we require an up-to-scale accurate estimate of camera pose and feature position to evaluate $r_{\mathcal{L}ik}$ for input to \cref{outlier rejection}. Therefore, we first solve the VI-BA without mono-depth (i.e., the pink rectangle portion of \cref{fig:Bundle Adjustment Factor Graph}). Finally after convergence of the depth-less cost-function, we add the depth constraints as detailed in this section, and solve \cref{ba_objective_function}.
\section{Experiments}
\label{sec:experiments}
\label{sec-experiments}
We perform two sets of experiments on the popular EuRoC dataset~\cite{euroc}, containing visual and inertial data from a micro air vehicle (MAV) along with accurate motion ground truth. To generate reliable correspondences for visual and mono-depth constraints, our front-end uses gyro measurements as a prior for frame-to-frame rotations following 2-pt RANSAC~\cite{2-pt-ransac}. We first exhaustively evaluate VIO initialization performance on the whole trajectory by running our initialization routine in windows sampled throughout each trajectory in the dataset, which is commonly done in a variety initialization works ~\cite{Inertial-only, analytical-solution, vertical-edge-vio-initialization}. Additionally, we also evaluate the effect of initialization on tracking performance by employing our method on a baseline similar to OpenVINS~\cite{openvins} in $10s$ time windows distributed uniformly across datasets. In both cases, we compare against ground truth poses captured by a VICON system present in the dataset. 

\subsection{Exhaustive Initialization Evaluation}
\label{exhaustive_eval}

Following prior related initialization works~\cite{Inertial-only, analytical-solution, vertical-edge-vio-initialization}, we exhaustively create VIO initialization events across the whole trajectory to evaluate performance across different motion and visual scenarios. For a fair comparison, we split each dataset into segments evenly and attempt to initialize all methods on the same set of segments. We collect poses from all successful initializations for the evaluation, \textbf{though note:} not all trials are successful due to internal validation steps of the respective algorithms and success does not necessarily mean that the initialization poses are \textbf{qualified} for tracking. Accuracy may be poor (measured by scale error or RMSE), in which case tracking may diverge. 

Our baseline method consists of a closed-form initialization~\cite{convex-initialization} followed by VI-BA~\cite{leutenegger2015keyframe} with only the VI-SFM portion of residuals present (pink rectangle in \cref{fig:Bundle Adjustment Factor Graph}). We also compare against the state-of-the-art VI-initialization method Inertial-only ~\cite{Inertial-only}, implementation of which is obtained from the open-sourced SLAM method~\cite{orb-slam3}. Given $N$ keyframes, Inertial-only uses up-to-scale visual odometry as the prior in a MAP framework to recover the metric scale, gravity vector, and IMU biases, followed by a VI-BA refinement step. Inertial-only's visual front-end performs RANSAC with PnP \cite{epnp}. 
 
We configured all three methods to operate on $10$Hz image streams following previous works~\cite{Inertial-only, analytical-solution, vins-mono-initialization}. We treat each image as a keyframe and use either $5$ or $10$ keyframes (KFs) for initialization. In the $5$KFs setting, we split datasets into $0.8s$ initialization windows evenly. For practical applications, faster initialization is preferred. So we specifically highlight a $5$KFs experiment to further exacerbate issues of insufficient baseline/motion, which are commonplace in deployment scenarios (e.g. MAVs, AR/VR). Other detailed experimental results for $10$KFs under $10Hz/4Hz$ settings (also studied in~\cite{Inertial-only}) are presented in the supplemental material.

We were able to generate $1078$, $1545$, $1547$, initialization trajectories respectively for Inertial-only, baseline, and our proposed method over all EuRoC datasets from $1680$ initialization attempts. The average initialization trajectory latency for the three methods were $0.592s$, $0.399s$, and $0.399s$ respectively. For our $10$KFs setting, we split datasets into $1.6s$ windows. We generated $571$, $809$, $815$ initialization trajectories for the three methods with an average trajectory latency of $1.367$, $0.897$ and $0.897$ from $839$ initialization attempts. Since Inertial-only uses visual odometry as the prior, to better align with the resulting expectations across different methods, we rejected those trajectories with poor resulting reprojection error of each visual constraint for the baseline and our proposed method. We observed that Inertial-only had longer initialization latency and typically led to fewer successful initializations because it requires mean trajectory acceleration larger than $0.5\%$ of gravity (${\bar{|| a ||}} > 0.005G$) as stated in \cite{Inertial-only}.

To measure trajectory accuracy, we perform a $Sim(3)$ alignment against the ground truth trajectory to get scale error and position RMSE for each initialization.  Since the global frames of the IMU sensor should be gravity-aligned, the gravity RMSE (in degrees) is computed from the global $z$ axis angular deviation in the IMU frame. Following past work~\cite{Inertial-only}, we omit scale errors when the mean trajectory acceleration ${\bar{|| a ||}}<0.005G$, however gravity and position RMSE are still reported. Finally, we also empirically compute the condition number of the problem hessian in the most challenging of sequences (mean acceleration ${\bar{|| a ||}}<0.005G$) to evaluate problem conditioning with the added mono-depth constraints. We present our aggregated results for the $5$KFs setting in \cref{benchmark result table}. We significantly outperform state-of-the-art Inertial-only in all metrics, achieving on average a $43\%$ reduction in scale error, $61\%$ reduction in position RMSE, and $21\%$ reduction in gravity RMSE for the challenging $5KF$ setting at an initialization latency of $0.4s$. Furthermore, our formulation leads to a lower condition number compared to the baseline, indicating improved problem conditioning. 

\begin{table*}[h]
\centering
\resizebox{\textwidth}{!}{
\begin{tabular}{@{}c c c c c c c c c c c c@{}} \toprule
   & \multicolumn{3}{c}{\textbf{Scale Error (\%)}} & \multicolumn{3}{c}{\textbf{Position RMSE}} &
 \multicolumn{3}{c}{\textbf{Gravity RMSE}} & \multicolumn{2}{c}{\textbf{$\log$(Condition Num)}} \\
   & \multicolumn{3}{c}{${\bar{|| a ||}} > 0.005G$} & \multicolumn{3}{c}{\textbf{(meters)}} &
 \multicolumn{3}{c}{\textbf{(degrees)}} & \multicolumn{2}{c}{${\bar{|| a ||}} < 0.005G$} \\
 \cmidrule(lr){2-4}\cmidrule(lr){5-7}\cmidrule(lr){8-10}\cmidrule(lr){11-12}
 \textbf{Dataset} & \textbf{Inertial-only} & \textbf{Baseline} & \textbf{Ours} & \textbf{Inertial-only} & \textbf{Baseline} & \textbf{Ours} & \textbf{Inertial-only} & \textbf{Baseline} & \textbf{Ours} & \textbf{Baseline} & \textbf{Ours} \\\midrule
 
mh\_01 & 41.34 & 43.65 & \textbf{31.11}  & 0.047  &  0.035 &  \textbf{0.025} & \textbf{1.38} &  2.43 &  1.82  &  13.97  & \textbf{13.16} \\
mh\_02 & 38.80  & 41.41 & \textbf{34.98} & 0.048  &  0.033 &  \textbf{0.026} & \textbf{1.33} &  2.04 &  1.81  &  13.31  & \textbf{12.50}  \\
mh\_03 & 57.44 & 59.09 & \textbf{34.65}  & 0.145  &  0.091 &  \textbf{0.055} & 3.09 &  3.73 &  \textbf{2.89}  &  13.83  & \textbf{12.73} \\
mh\_04 & 74.29 & 56.26 & \textbf{48.40}  & 0.179  &  0.090 &  \textbf{0.075} & 2.38 &  2.69 &  \textbf{2.31}  &  13.42  & \textbf{11.27} \\
mh\_05 & 70.35 & 54.64 & \textbf{44.52}  & 0.145  &  0.078 &  \textbf{0.063} & \textbf{2.13} &  2.77 &  2.30   &  13.66  & \textbf{12.51} \\
v1\_01 & 55.44 & 54.25 & \textbf{25.59}  & 0.056  &  0.038 &  \textbf{0.021} & 3.47 &  3.73 &  \textbf{3.36}  &  12.93  & \textbf{11.43} \\
v1\_02 & 56.86 & 45.12 & \textbf{26.12}  & 0.106  &  0.069 &  \textbf{0.038} & 3.77 &  3.86 &  \textbf{2.44}  &  13.26  & \textbf{11.67} \\
v1\_03 & 56.93 & 38.55 & \textbf{20.01}  & 0.097  &  0.048 &  \textbf{0.025} & 5.36 &  3.59 &  \textbf{2.37}  &  12.62  & \textbf{12.03} \\
v2\_01 & 42.40  & 40.84 & \textbf{23.51} & 0.035  &  0.026 &  \textbf{0.015} & 1.49 &  1.78 &  \textbf{1.35}  &  13.45  & \textbf{12.84} \\
v2\_02 & 41.27 & 34.31 & \textbf{19.33}  & 0.035  &  0.026 &  \textbf{0.015} & 2.92 &  2.66 &  \textbf{1.96}  &  \textbf{12.20}   & 12.27 \\
v2\_03 & 59.64 & 36.42 & \textbf{27.87}  & 0.116  &  0.044 &  \textbf{0.033} & 4.10  &  2.81 &  \textbf{2.24}  &  13.30   & \textbf{11.17} \\
\midrule
Mean   & 54.07 & 45.87 & \textbf{30.55}  & 0.092  &  0.053 &  \textbf{0.036} &  2.86 &  2.92 &  \textbf{2.26}  & 13.27 & \textbf{12.14} \\

\bottomrule
\end{tabular}
}

\caption{Exhaustive initialization benchmark results per dataset from Inertial-only, our baseline, and our proposed method using $5$ KFs with $10Hz$ image data. For each metric, lower is better.}
\label{benchmark result table}
\end{table*}

To demonstrate the importance of the scale/shift priors, edge weighting, and outlier rejection introduced in this work, we present results of an ablation study in \cref{ablation study}. This study shows each component significantly improves the overall performance of the system.

\begin{table*}[h]
\centering
\resizebox{0.8\textwidth}{!}{
\begin{tabular}{@{}c c c c c c@{}} \toprule
\multirow{2}{*}{\textbf{Metrics}} & \multirow{2}{*}{\textbf{Ours}} & \textbf{Ours w/o} & \textbf{Ours w/o} & \textbf{Ours w/o} & \textbf{Ours w/o} \\
& & \textbf{Prior} & \textbf{Weight} & \textbf{Outlier Rejection} & \textbf{Everything}  \\
\midrule
\midrule

\multicolumn{1}{l}{Scale Error (\%) ${\bar{|| a ||}} > 0.005G$}  &  \textbf{31.23} & 35.47 & 34.22 & 36.59 & 37.55 \\
\cmidrule(lr){1-1}\cmidrule(lr){2-6}
\multicolumn{1}{l}{Position RMSE (meters)}  & \textbf{0.036} & 0.041 & 0.041 & 0.039 & 0.044\\
\cmidrule(lr){1-1}\cmidrule(lr){2-6}
\multicolumn{1}{l}{Gravity RMSE (degrees)} & \textbf{2.26} & 2.53 & 2.46 & 2.46 & 2.57\\
\cmidrule(lr){1-1}\cmidrule(lr){2-6}
\multicolumn{1}{l}{\textbf{$\log$(Condition Num)} ${\bar{|| a ||}} < 0.005G$} & \textbf{12.14} & 13.24 & 13.23 & 13.18 & 13.49 \\
\bottomrule
\end{tabular}
}
\caption{Aggregated exhaustive initialization benchmark ablation study of our proposed method using $5$ KFs with $10Hz$ image data for all EuRoC datasets. For each metric, lower is better.}
\label{ablation study}
\end{table*}

\begin{figure}[t]
    \centering
    \includegraphics[width=0.8\linewidth]{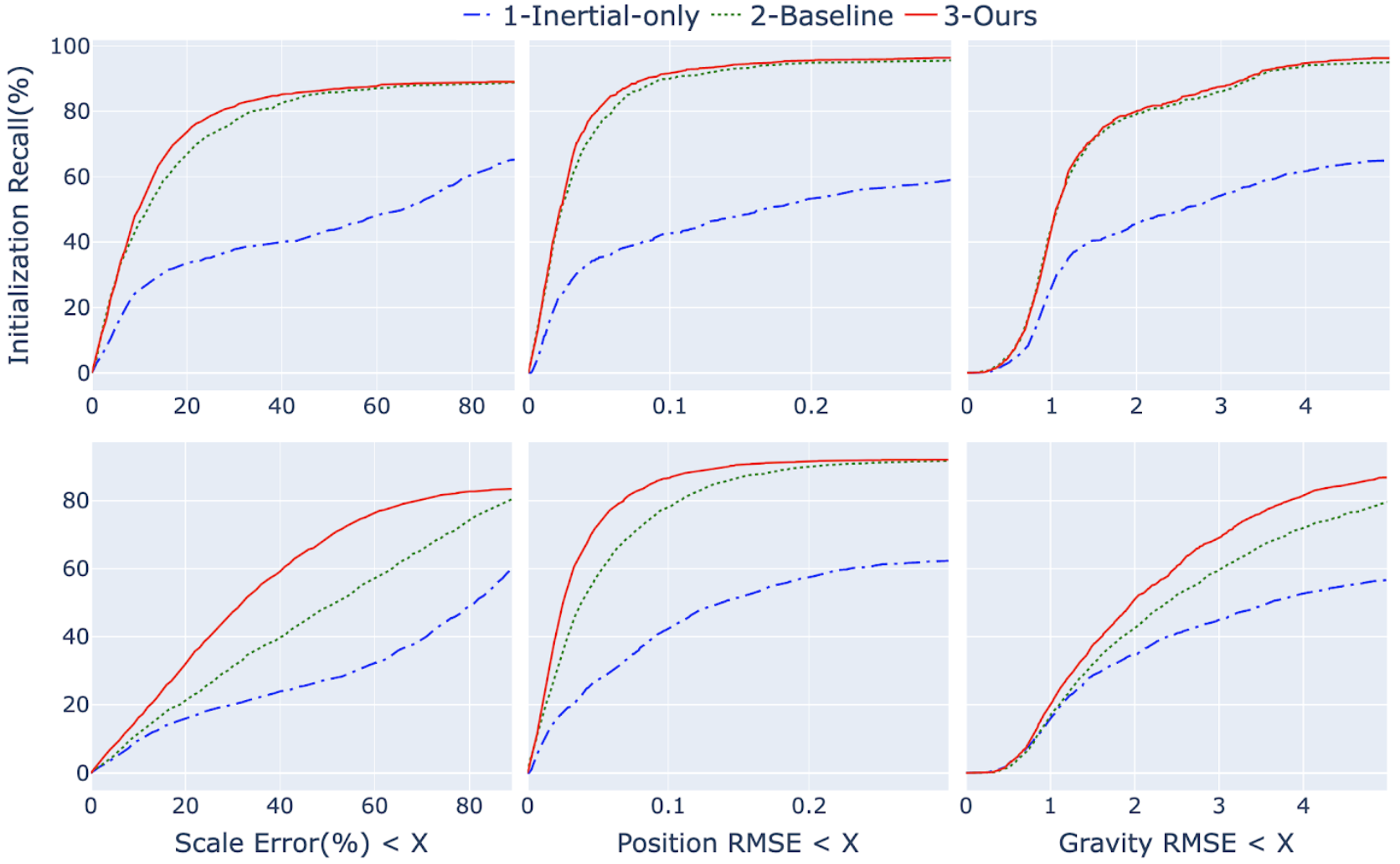}
    \caption{Cumulative distribution plots for primary error metrics. \textbf{First row}: Results with 10 keyframes. \textbf{Second row}: Results with 5 keyframes. For each plot, the $X$ axis denotes a threshold for error metric and the $Y$ axis shows the fraction of initialization attempts with the respective error metric smaller than the threshold on the $X$ axis. \textbf{Note:} 1) Improved gains in the $5$KF (i.e. less motion) setting where mono-depth residuals show greater impact. 2) Recall doesn't converge to 100\% due to initialization failures among attempts.}
    \label{fig:percentage-error}
\end{figure}

\begin{figure}[h]
    \centering
    \includegraphics[width=1\linewidth]{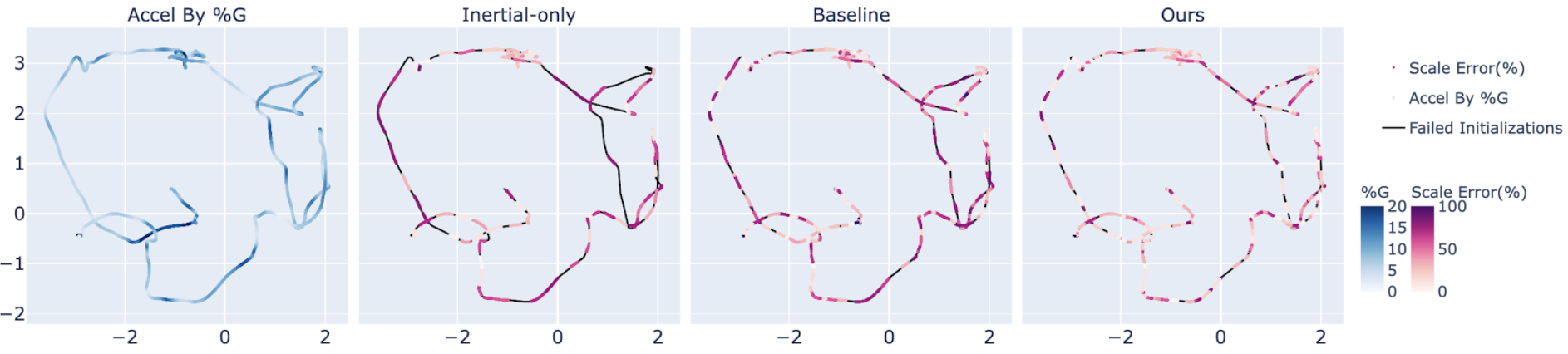}
    \caption{Acceleration and scale error visualizations for the v2\_01 dataset (best viewed in color). \textbf{Left:} Trajectory colored by acceleration magnitude as \%G (lighter indicates low acceleration). \textbf{Right:} Segments of poses colored by scale error magnitude for each initialization window in the dataset (lighter is better). Segments colored black indicate failed initializations for the respective methods. We outperform  other methods over the entire trajectory on scale error, especially in low acceleration regions, e.g, left side of the plot, where our method performs significantly better.}
    \label{fig:colormap-traj-vis}
\end{figure}

In \cref{fig:percentage-error}, we plot the cumulative distributions for the metrics above for both the $10$KFs (top) and $5$KFs (bottom) settings. We can see that while we do better than the baseline and Inertial-only in the $10$KFs setting, the gains are greater in the more challenging $5$ KFs setting with low-excitation, highlighting the benefit of the mono-depth residuals. In order to gain insights into where our method outperforms others, we visualize a dataset with trajectory color coded by acceleration magnitude and scale error for the various methods in \cref{fig:colormap-traj-vis}. We outperform both Inertial-only and the baseline almost across the whole trajectory but more specifically so in low acceleration regions which are traditionally the hardest for classical VIO initialization methods. This further validates our hypothesis that the added mono-depth constraints condition the system better with direct (up to scale/shift) depth measurement priors in low-excitation scenarios, which is critical for today's practical applications of VIO. 


\subsection{Visual-inertial Odometry Evaluation}
\begin{table}[b]
\centering
\resizebox{0.5\textwidth}{!}{
\begin{tabular}{@{}c c c c c c c@{}} \toprule
   & \multicolumn{3}{c}{\textbf{Position RMSE (m)}} & \multicolumn{3}{c}{\textbf{Gravity RMSE (deg)}} \\
    \cmidrule(lr){2-4}\cmidrule(lr){5-7}
  \textbf{Dataset} & \textbf{Baseline} & \textbf{Ours} & \textbf{Diff(\%)} & \textbf{Baseline} & \textbf{Ours} & \textbf{Diff(\%)} \\
    \midrule
mh\_01 &  1.560 & \textbf{0.543} & -65.19 & 2.21  & \textbf{1.55}  & -29.86 \\
mh\_02 &  0.604 & \textbf{0.071} & -88.24 & 1.65  & \textbf{1.31}  & -20.60 \\
mh\_03 &  2.466 & \textbf{1.299} & -47.32 & 2.88  & \textbf{2.29}  & -20.48 \\
mh\_04 &  0.526 & \textbf{0.124} & -76.42 & 2.01  & \textbf{1.01}  & -49.75 \\
mh\_05 &  3.204 & \textbf{0.910} & -71.59 & 3.44  & \textbf{1.88}  & -45.34 \\
v1\_01 &  3.438 & \textbf{0.082} & -97.61 & 4.66  & \textbf{2.69}  & -42.27 \\
v1\_02 &  2.846 & \textbf{0.097} & -96.59 & 3.57  & \textbf{1.22}  & -65.82 \\
v1\_03 &  2.649 & \textbf{0.059} & -97.77 & 3.19  & \textbf{1.28}  & -59.87 \\
v2\_01 &  1.824 & \textbf{0.046} & -97.47 & 2.19  & \textbf{1.08}  & -50.68 \\
v2\_02 &  2.615 & \textbf{0.060} & -97.70 & 3.42  & \textbf{1.25}  & -63.45 \\
v2\_03 &  2.939 & \textbf{0.567} & -80.70 & 3.99  & \textbf{2.06}  & -48.37 \\ 
\midrule
Mean   &  2.243 & \textbf{0.351} & -84.35 & 3.02  & \textbf{1.61} & -46.68 \\
\bottomrule
\end{tabular}
}
\caption{Visual-inertial odometry benchmark results over all EuRoC datasets with and without mono-depth constraints used in initialization. VIO runs at $10Hz$ and is initialized with $5$KFs.}
\label{visual-inertial odometry benchmark result}
\end{table}

To better illustrate our method's in-the-wild applicability, we conduct experiments quantifying the impact of our method when used in-the-loop with odometry. Considering the additional challenge of $5$KFs initialization, we focus our experiments there instead of typical $10$KFs~\cite{Inertial-only} and evaluate the accuracy of final tracking trajectories. The evaluation is performed with a baseline similar to OpenVINS~\cite{openvins}, which is a state-of-the-art VIO system commonly used in compute-limited use-cases (e.g, mobile AR/VR, drones). Similar to \cref{exhaustive_eval}, we create initialization events periodically but evaluate the tracking trajectories instead. We split the datasets evenly into $10s$ segments and initialize and perform VIO using the same $10s$ of information for both methods.

As in \cref{exhaustive_eval}, our baseline is tracking initialized with VI-SFM only. We generated a total of $142$ trajectories using our protocol over all EuRoC datasets for each method and report aggregated position and gravity RMSE for each dataset. The aggregated results are shown in \cref{visual-inertial odometry benchmark result} where we see an $84\%$ improvement in position RMSE and $46\%$ improvement in gravity RMSE over the baseline method. This suggests a significant expected improvement in downstream uses of odometry, such as rendering virtual content, depth estimation, or navigation.

\textbf{Computation Cost.} We ran our system on a Pixel4XL mobile phone using only CPU cores. The computation cost (in milliseconds) for different initialization modules is shown in \cref{computation-cost}. The closed-form initialization problem is solved using Eigen~\cite{eigen} and the subsequent VI-BA is solved with the Ceres Solver~\cite{ceres-solver} using Levenberg–Marquardt. We run ML inference on the CPU in its own thread and hence achieve real-time performance (within $100ms$ for the $10Hz$ configuration) on a mobile phone. While we do observe that adding depth constraints increases the computational cost of the VI-SFM problem, we still improve in terms of overall initialization speed by producing a satisfactory solution with only $5$KFs (\textbf{0.5s of data}) as opposed to $10$KFs typically required by the baseline and Inertial-only.
\begin{table}[t]
\centering
\resizebox{0.8\textwidth}{!}{
\begin{tabular}{@{}c | c | c | c@{}} \toprule
  \textbf{Mono depth} & \textbf{Closed-form Initialization} & \textbf{VI-BA Solver (baseline)} & \textbf{VI-BA Solver (ours)}\\
 \midrule
  71.64 & 0.73 & 16.2 & 39.8 \\
 \bottomrule
\end{tabular}
}
\caption{Computation duration of key modules in milliseconds.}
\label{computation-cost}
\end{table}

\section{Conclusion }
In this paper, we introduced a novel VIO initialization method leveraging learned monocular depth. We integrated the learned depth estimates, with alignment parameters, into a classical VI-SFM formulation. Through the learned image priors, our method gains significant robustness to typical degenerate motion configurations for VI-SFM, such as low parallax and low excitation (near-zero) acceleration. This method only requires a lightweight ML model and additional residuals (with associated states) to be added to a standard pipeline and does not significantly impact runtime, enabling application on mobile devices. Our experiments demonstrated significant improvements to accuracy, problem conditioning, and robustness relative to the state-of-the-art, even when significantly reducing the number of keyframes used and exacerbating the problem of low excitation. Our method could serve as a straightforward upgrade for most traditional pipelines. There are several key limitations and directions for future work to call out: 
\begin{itemize}
    \setlength\itemsep{0.1em}
    \item We do not claim any direct upgrades to VI system observability. While the use of a prior on scale and shift and the training of the mono-depth network (assuming scale and shift being $1$ and $0$) may provide some direct scale information, our work's primary contribution is to problem conditioning and behaviour under \textit{limited} motion, not \textit{zero} motion.
    \item Mono-depth has generalization limitations due to biases in its training data, learning scheme, and model structure. It is crucial to note that we \textbf{did not re-train} our network for EuRoC. It was used \textit{off the shelf} after training on general imagery which are very different from EuRoC. With a network trained specifically for the problem domain (or optimized in the loop at test time per initialization window) we expect an even greater improvement.
\end{itemize}

\textbf{Acknowledgements.} We thank Josh Hernandez and Maksym Dzitsiuk for their support in developing our real-time system implementation.

\clearpage
%
%
\bibliographystyle{splncs04}
\bibliography{egbib}
\end{sloppypar}
\end{document}